# A Simple Insight into Iterative Belief Propagation's Success

Rina Dechter and Robert Mateescu
School of Information and Computer Science
University of California, Irvine, CA 92697-3425
{dechter, mateescu}@ics.uci.edu

## Abstract

In non-ergodic belief networks the posterior belief of many queries given evidence may become zero. The paper shows that when belief propagation is applied iteratively over arbitrary networks (the so called, iterative or loopy belief propagation (IBP)) it is identical to an arc-consistency algorithm relative to zero-belief queries (namely assessing zero posterior probabilities). This implies that zero-belief conclusions derived by belief propagation converge and are sound. More importantly, it suggests that the inference power of IBP is as strong and as weak as that of arc-consistency. This allows the synthesis of belief networks for which belief propagation is useless on one hand, and focuses the investigation on classes of belief networks for which belief propagation may be zero-complete. Finally, we show empirically that IBP's accuracy is correlated with extreme probabilities, therefore explaining its success over coding applications.

## 1 INTRODUCTION AND MOTIVATION

The paper proves that zero-belief conclusions made by iterative belief propagation are correct. The proof is done using a transformation of the belief propagation algorithm to an arc-consistency algorithm in constraint networks.

The belief propagation algorithm is a distributed algorithm that computes posterior beliefs for tree-structured Bayesian networks (poly-trees) [Pearl1988]. However, in recent years it was shown to work surprisingly well in many applications involving networks with loops, including turbo codes, when applied iteratively [McEliece et al.1997].

The success of Iterative Belief Propagation (IBP) inspired extensions into a general class of *Generalized belief propagation* [Yedidia et al.2001] algorithms that try to improve on its performance. A class of such algorithms called *iterative join-graph propagation*, IJGP(i) [Dechter et al.2002], extends IBP into anytime algorithms. Algorithm IJGP applies the belief propagation idea to clusters of functions that form a join-graph rather than to single functions, and its parameter $i$ allows the user to control the tradeoff between complexity and accuracy. When the join-graph is a tree the algorithm is exact. Empirical evaluation of this class of algorithms showed that IJGP improves substantially over IBP and provides anytime performance.

While there is still very little understanding as to why and when IBP works well, some recent investigation shows that when IBP converges, it converges to a stationary point of the Bethe energy, thus making connections to approximation algorithms developed in statistical physics and to variational approaches to approximate inference [Welling and Teh2001, Yedidia et al.2001]. However, these approaches do not explain why IBP is successful where it is, and do not allow any performance guarantees on accuracy.

This paper makes some simple observations that may shed light on IBP's behavior, and on the more general class of IJGP algorithms. Zero-beliefs are variable-value pairs that have zero conditional probability given the evidence. We show that: if a value of a variable is assessed as having zero-belief in any iteration of IBP, it remains a zero-belief in all subsequent iterations; that IBP finitely converges relative to its set of zero-beliefs; and, most importantly that the set of zero-belief decided by any of the iterative belief propagation methods is sound. Namely any zero-belief determined by IBP corresponds to a true zero conditional probability relative to the given probability distribution expressed by the Bayesian network.

While each of these claims can be proved directly, our approach is to associate a belief network with a constraint network and show a correspondence between IBP applied to the belief network and an arc-consistency algorithm applied to the corresponding constraint network. Since arc-consistency algorithms are well understood this correspondence not only proves right away the targeted claims, but may provide additional insight into the behavior of IBP and IJGP. In particular, not only it immediately justifies the iterative application of belief propagation algorithms on one hand, but it also illuminates its "distance" from being complete, on the other.

Section 2 provides preliminaries, section 3 describes the class of dual join-graphs and defines IBP as an instance of propagation on dual join-graphs, section 4 relates the belief network to a constraint network and IBP to arc-consistency, and describes classes of strong and weak inference power for IBP, section 5 provides empirical evaluation and section 6 concludes.

## 2 PRELIMINARIES AND BACKGROUND

**Belief networks.** *Belief networks* provide a formalism for reasoning about partial beliefs under conditions of uncertainty. A belief network is defined by a directed acyclic graph over nodes representing random variables. Formally, it is a quadruple $BN = <X, D, G, P>$ (also abbreviated $<G, P>$ when $X$ and $D$ are clear) where $X = \{X_1, ..., X_n\}$ is a set of random variables, $D = \{D_1, ..., D_n\}$ is the set of the corresponding domains, $G$ is a directed acyclic graph over $X$ and $P = \{p_1, ..., p_n\}$, where $p_i = P(X_i|pa_i)$ ($pa_i$ are the parents of $X_i$ in $G$) denote conditional probability tables (CPTs), each defined on a variable and its parent set. The belief network represents a probability distribution over $X$ having the product form $P(x_1, ..., x_n) = \prod_{i=1}^{n} P(x_i|x_{pa_i})$, where an assignment $(X_1 = x_1, ..., X_n = x_n)$ is abbreviated to $x = (x_1, ..., x_n)$ and where $x_s$ denotes the restriction of a tuple $x$ to the subset of variables $S$. An evidence set $e$ is an instantiated subset of variables. We use upper case letters for variables and nodes in a graph and lower case letters for values in variables' domains. Given a function $f$, we denote the set of arguments of $f$ by $scope(f)$. The family of $X_i$, denoted by $F_i$, includes $X_i$ and its parent variables.

**Belief updating.** The *belief updating* problem defined over a belief network (also referred to as *probabilistic inference*) is the task of computing the posterior probability $P(Y|e)$ of *query* nodes $Y \subseteq X$ given evidence $e$. We will focus on two cases: 1) when $Y$ consists of a single variable $X_i$; namely on computing $Bel(X_i) = P(X_i = x|e)$, $\forall X_i \in X$, $\forall x \in D_i$; 2) when $Y$ consists of the scope of an original CPT; that is, we compute $Bel(F_i) = P(F_i = t|e)$, $\forall F_i\ family\ in\ \mathcal{B}$, $\forall t \in \times_{X_i \in F_i} D_i$.

**Constraint networks.** A *constraint network* $\mathcal{R} = (X, D, C)$ is defined over a set of variables $X = \{X_1, ..., X_n\}$, their respective domains of values $D = \{D_1, ..., D_n\}$ and a set of constraints $C = \{C_1, ..., C_t\}$. Each constraint is a pair $C_i = (S_i, R_i)$, where $S_i \subseteq X$ is the scope of the relation $R_i$, and $R_i$ defines the allowed combinations of values. In a *binary constraint network* each constraint, denoted $R_{ij}$, is defined over pairs of variables $X_i$ and $X_j$. The primary query over constraint networks is to determine if there exists a solution, namely an assignment $x = (x_1, ..., x_n)$ to all the variables that satisfies all the constraints (i.e. $\forall i, x_{S_i} \in R_i$), and if so, to find one. A constraint networks can be associated with a constraint graph where each node represents a variable, and any two variables appearing in the same constraint's scope are connected. We say that a constraint network $\mathcal{R}$ represents its set of all solutions $sol(\mathcal{R})$.

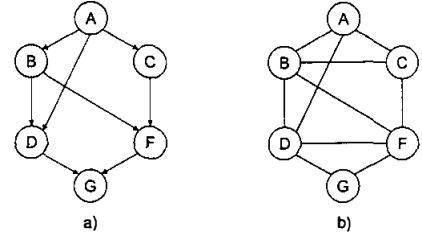

Figure 1: a) A belief network: $P(g, f, d, c, b, a) = P(g|f, d) \cdot P(f|c, b) \cdot P(d|b, a) \cdot P(b|a) \cdot P(c|a) \cdot P(a)$; b) A constraint network with relations having the same scopes;

**Example 2.1** *Figure 1a gives an example of a belief network over 6 variables and Figure 1b shows a constraint network over the same set of variables, with relations having the same scopes as the functions of 1a.*

**DEFINITION 2.1 (dual graphs)** *Given a set of functions $F = \{f_1, ..., f_l\}$ over scopes $S_1, ..., S_l$, the dual graph of $F$ is a graph $DG = (V, E, L)$ that associates a node with each function, namely $V = F$ and an arc connects any two nodes whose scope share a variable, $E = \{(f_i, f_j)|S_i \cap S_j \neq \phi\}$. $L$ is a set of labels for the arcs, each arc being labeled by the shared variables of its nodes, $L = \{l_{ij} = S_i \cap S_j|(i, j) \in E\}$.*

The definition of a dual graph is applicable both to Bayesian networks and to constraint networks. It is known that the dual graph of a constraint network transforms any non-binary network into a binary one, where the domains of the variables are the allowed tuples in each relation and the constraints of the dual problem force equality over shared variables labeling the arcs [Dechter1992].

**Constraint propagation** algorithms is a class of polynomial time algorithms that are at the center of constraint processing techniques. They were investigated extensively in the past three decades and the most well known versions are arc-, path-, and i-consistency [Dechter1992].

**DEFINITION 2.2 (arc-consistency)** *[Mackworth1977] Given a binary constraint network $(X, D, C)$, the network is arc-consistent iff for every binary constraint $R_{ij} \in C$, every value $v \in D_i$ has a value $u \in D_j$ s.t. $(v, u) \in R_{ij}$.*

When a binary constraint network is not arc-consistent, arc-consistency algorithms can enforce arc-consistency. The algorithms remove values from the domains of the variables that violate arc-consistency until an arc-consistent network is generated. A variety of improved performance arc-consistency algorithms were developed over the years, however for the sake of this paper we will consider a non-optimal distributed version, which we call *distributed arc-consistency*.

DEFINITION 2.3 (**distributed arc-consistency**) *The algorithm is a message passing algorithm. Each node maintains a current set of viable values $D_i$. Let $ne(i)$ be the set of neighbors of $X_i$ in the constraint graph. Every node $X_i$ sends a message to any node $X_j \in ne(i)$, which consists of the values in $X_j$'s domain that are consistent with the current $D_i$, relative to the constraint that they share. Namely, the message that $X_i$ sends to $X_j$, denoted by $D_i^j$, is:*

$$D_i^j \leftarrow \pi_j(R_{ji} \bowtie D_i) \qquad (1)$$

*(where, join ($\bowtie$) and project ($\pi$) are the usual relational operators) and in addition node i computes:*

$$D_i \leftarrow D_i \cap (\bowtie_{k \in ne(i)} D_k^i) \qquad (2)$$

Clearly the algorithm can be synchronized into iterations, where in each iteration every node computes its current domain based on all the messages received so far from its neighbors (eq. 2), and sends a new message to each neighbor (eq. 1). Alternatively, equations 1 and 2 can be combined. The message $X_i$ sends to $X_j$ is:

$$D_i^j \leftarrow \pi_j(R_{ji} \bowtie D_i \bowtie_{k \in ne(i)} D_k^i) \qquad (3)$$

The above distributed arc-consistency algorithm can be applied to the dual problem of any non-binary constraint network as well. This is accomplished by the following rule applied by each node in the dual graph. We call the algorithm distributed relational arc-consistency.

DEFINITION 2.4 (**distributed relational arc-consistency; DR-AC**) *Let $R_i$ and $R_j$ be two constraints sharing scopes, whose arc in the dual graph is labeled by $l_{ij}$. The message $R_i$ sends to $R_j$ denoted $h_i^j$ is defined by:*

$$h_i^j \leftarrow \pi_{l_{ij}}(R_i \bowtie (\bowtie_{k \in ne(i)} h_k^i)) \qquad (4)$$

**Example 2.2** *Figure 2 describes part of the execution of DR-AC for a problem inspired by graph coloring, on the graph of figure 1b. All variables have the same domain, $\{1,2,3\}$, except for C which is 2, and G which is 3. Arcs in figure 1b correspond to* not equal *constraints. The dual graph of this problem is given in figure 2, and each table shows the initial constraints (there are unary, binary and ternary constraints). To initialize the algorithm, the first messages sent out by each node are universal relations over the labels. For this example, DR-AC actually solves the problem and finds the unique solution A=1, B=3, C=2, D=2, F=1, G=3.*

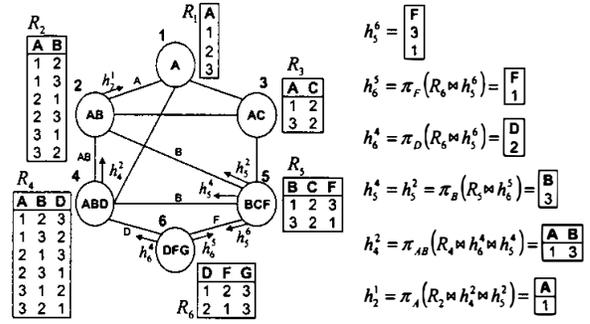

Figure 2: Part of the execution of DR-AC algorithm;

**Proposition 1** *Distributed relational arc-consistency converges after $O(t \cdot r)$ iterations to the largest arc-consistent network that is equivalent to the original network, where t bounds the number of tuples in each constraint and r is the number of constraints.*

**Proposition 2 (complexity)** *The complexity of distributed arc-consistency is $O(r^2 t^2 \log t)$.*

## 3 ITERATIVE BELIEF PROPAGATION OVER DUAL JOIN-GRAPHS

Iterative belief propagation (IBP) is an iterative application of Pearl's algorithm that was defined for poly-trees [Pearl1988]. Since it is a distributed algorithm, it is well defined for any network. In this section we will present IBP as an instance of join-graph propagation over variants of the *dual graph*.

Consider a Bayesian network $\mathcal{B} = <X, D, G, P>$. As defined earlier, the *dual graph* $\mathcal{D}_\mathcal{G}$ of the Belief network $\mathcal{B}$, is an arc-labeled graph defined over the CPTs as its functions. Namely, it has a node for each CPT and a labeled arc connecting any two nodes that share a variable in the CPT's scope. The arcs are labeled by the shared variables. A *dual join-graph* is a labeled arc subgraph of $\mathcal{D}_\mathcal{G}$ whose arc labels are subsets of the labels of $\mathcal{D}_\mathcal{G}$ such that the *running intersection property*, also called *connectedness property*, is satisfied. The running intersection property requires that any two nodes that share a variable in the dual join-graph be connected by a path of arcs whose labels contain the shared variable. Clearly the dual graph itself is a dual join-graph. An *arc-minimal* dual join-graph is a dual join-graph for which none of the labels can be further reduced while maintaining the connectedness property.

Interestingly, there are many dual join-graphs of the same dual graph and many of them are arc-minimal. We define Iterative Belief Propagation on a dual join-graph. Each node sends a message over an arc whose scope is identical to the label on that arc. Since Pearl's algorithm sends messages whose scopes are singleton variables only, we highlight arc-minimal singleton dual join-graph. One such

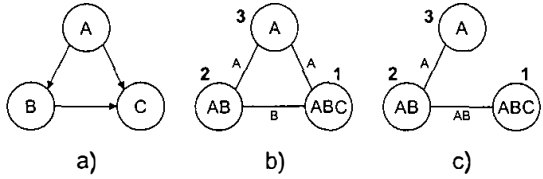

Figure 3: a) A belief network; b) A dual join-graph with singleton labels; c) A dual join-graph which is a join-tree;

graph can be constructed directly from the graph of the Bayesian network, labeling each arc with the parent variable. It can be shown that:

**Proposition 3** *The dual graph of any Bayesian network has an arc-minimal dual join-graph where each arc is labeled by a single variable.*

**Example 3.1** *Consider the belief network on 3 variables $A, B, C$ with CPTs 1.$P(C|A, B)$, 2.$P(B|A)$ and 3.$P(A)$, given in Figure 3a. Figure 3b shows a dual graph with singleton labels on the arcs. Figure 3c shows a dual graph which is a join tree, on which belief propagation can solve the problem exactly in one iteration (two passes up and down the tree).*

We will next present IBP algorithm that is applicable to any dual join-graph (Figure 4). The algorithm is a special case of IJGP introduced in [Dechter et al.2002]. It is easy to see that one iteration of IBP is time and space linear in the size of the belief network, and when IBP is applied to the singleton labeled dual graph it coincides with Pearl's belief propagation applied directly to the acyclic graph representation. For space reasons, we do not include the proof here. Also, when the dual join-graph is a tree IBP converges after one iteration (two passes, up and down the tree) to the exact beliefs.

## 4 FLATTENING THE BAYESIAN NETWORK

Given a belief network $\mathcal{B}$ we will now define a flattening of the Bayesian network into a constraint network called $flat(\mathcal{B})$ where all the zero entries in the CPTs are removed from the corresponding relation. $flat(B)$ is a constraint network defined over the same set of variables and has the same set of domain values as $\mathcal{B}$. Formally, for every $X_i$ and its CPT $P(X_i|pa_i) \in \mathcal{B}$ we define a constraint $R_{F_i}$ over the family of $X_i$, $F_i = \{X_i\} \cup pa_i$ as follows: for every assignment $x = (x_i, x_{pa_i})$ to $F_i$,

$$(x_i, x_{pa_i}) \in R_{F_i} \quad iff \quad P(x_i|x_{pa_i}) > 0.$$

The evidence set $e = \{e_1, ..., e_r\}$ is mapped into unary constraints that assign the corresponding values to the evidence variables.

---

**Algorithm IBP**
**Input:** An arc-labeled dual join-graph $DJ = (V, E, L)$ for a Bayesian network $BN = <X, D, G, P>$. Evidence $e$.
**Output:** An augmented graph whose nodes include the original CPTs and the messages received from neighbors. Approximations of $P(X_i|e)$, $\forall X_i \in X$. Approximations of $P(F_i|e)$, $\forall F_i \in \mathcal{B}$.
Denote by: $h_u^v$ the message from $u$ to $v$; $ne(u)$ the neighbors of $u$ in $V$; $ne_v(u) = ne(u) - \{v\}$; $l_{uv}$ the label of $(u, v) \in E$; $elim(u, v) = scope(u) - scope(v)$.
• **One iteration of IBP**
For every node $u$ in $DJ$ in a topological order and back, do:
1. **Process observed variables**
Assign relevant evidence to the each $p_i$ and remove the relevant variables from the labeled arcs.
2. **Compute and send to $v$ the function:**

$$h_u^v = \sum_{elim(u,v)} (p_u \cdot \prod_{\{h_i^u, i \in ne_v(u)\}} h_i^u)$$

**Endfor**
• **Compute approximations of** $P(F_i|e)$, $P(X_i|e)$:
For every $X_i \in X$ let $u$ be the vertex of family $F_i$ in $DJ$,
$P(F_i|e) = \alpha(\prod_{h_i^u, u \in ne(i)} h_i^u) \cdot p_u;$
$P(X_i|e) = \alpha \sum_{scope(u) - \{X_i\}} P(F_i|e).$

Figure 4: Algorithm Iterative Belief Propagation;

**THEOREM 4.1** *Given a belief network $\mathcal{B}$ and evidence $e$, for any tuple $t$: $P_\mathcal{B}(t|e) > 0 \Leftrightarrow t \in sol(flat(B, e))$.*

**Proof.** $P_\mathcal{B}(t|e) > 0 \Leftrightarrow \Pi_i P(x_i|x_{pa_i})|_t > 0 \Leftrightarrow \forall i, P(x_i|x_{pa_i})|_t > 0 \Leftrightarrow \forall i, (x_i, x_{pa_i})|_t \in R_{F_i} \Leftrightarrow t \in sol(flat(B, e))$, where $|_t$ is the restriction to $t$. □

We next define an algorithm dependent notion of zero tuples.

**DEFINITION 4.1 (IBP-zero)** *Given a CPT $P(X_i|pa_i)$, an assignment $x = (x_i, x_{pa_i})$ to its family $F_i$ is IBP-zero if some iteration of IBP determines that $P(x_i|x_{pa_i}, e) = 0$.*

It is easy to see that when IBP is applied to a constraint network where sum and product are replaced by join and project, respectively, it becomes identical to distributed relational arc-consistency defined earlier. Therefore, a partial tuple is removed from a flat constraint by arc-consistency iff it is IBP-zero relative to the Bayesian network.

**THEOREM 4.2** *When IBP is applied in a particular variable ordering to a dual join-graph of a Bayesian network $\mathcal{B}$, its trace is identical, relative to zero-tuples generation, to that of DR-AC applied to the corresponding flat dual join-graph. Namely, taking a snapshot at identical steps, any IBP-zero tuple in the Bayesian network is a removed tuple in the corresponding step of DR-AC over the flat dual join-graph.*

**Proof.** It suffices to prove that the first iteration of IBP and DR-AC generates the same zero tuples and removed tuples,

respectively. We prove the claim by induction over the topological ordering that defines the order in which messages are sent in the corresponding dual graphs.

*Base case:* By the definition of the flat network, when algorithms IBP and DR-AC start, every zero probability tuple in one of the CPTs $P_{X_i}$ in the dual graph of the Bayesian network, becomes a removed tuple in the corresponding constraint $R_{F_i}$ in the dual graph of the flat network.

*Inductive step:* Suppose the claim is true after $n$ correspondent messages are sent in IBP and DR-AC. Suppose the $(n + 1)th$ message is scheduled to be the one from node $u$ to node $v$. Indexing messages by the name of the algorithm, in the dual graph of IBP, node $u$ contains $p_u$ and $h_{IBP_i^u}, i \in ne_v(u)$, and in the dual graph of DR-AC, node $u$ contains $R_u$ and $h_{DR\ AC_i^u}, i \in ne_v(u)$. By the inductive hypothesis, the zero tuples in $p_u$ and $h_{IBP_i^u}, i \in ne_v(u)$ are the removed tuples in $R_u$ and $h_{DR\ AC_i^u}, i \in ne_v(u)$, respectively. Therefore, the zero tuples in the product $(p_u \cdot (\prod_{i \in ne_v(u)}) h_i^u)$ correspond to the removed tuples in the join $(R_u \bowtie (\bowtie_{i \in ne_v(u)}) h_i^u)$. This proves that the zero tuples in the message of IBP
$$h_{IBP_u^v} = \sum_{elim(u,v)} (p_u \cdot (\prod_{i \in ne_v(u)}) h_i^u),$$ correspond to the removed tuples in the message of DR-AC
$$h_{DR\ AC_u^v} = \pi_{l_{uv}}(R_u \bowtie (\bowtie_{i \in ne_v(u)}) h_i^u).$$
The same argument can now be extended for every iteration of the algorithms. □

**Corollary 1** *Algorithm IBP zero-converges. Namely, its set of zero tuples does not change after $t \cdot r$ iterations.*

**Proof.** From Theorem 4.2 any IBP-zero is a no-good removed by arc-consistency over the flat network. Since arc-consistency converges, the claim follows. □

**THEOREM 4.3** *When IBP is applied to a dual join-graph of a Bayesian network, any tuple $t$ that is IBP-zero satisfies $P_B(t|e) = 0$.*

**Proof.** From Theorem 4.2 if a tuple $t$ is IBP zero, it is also removed from the corresponding relation by arc-consistency over $flat(\mathcal{B}, e)$. Therefore this tuple is a no-good of the network $flat(\mathcal{B}, e)$ and, from Theorem 4.1 it follows that $P_B(t|e) = 0$. □

### 4.1 ZEROS ARE SOUND FOR ANY IJGP

The results for IBP can be extended to the more general class of algorithms called *iterative join-graph propagation*, IJGP [Dechter et al. 2002]. IJGP can be viewed as a generalized belief propagation algorithm and was shown to benefit both from the virtues of iterative algorithms and from the anytime characteristics of bounded inference provided by mini-buckets schemes.

The message-passing of IJGP is identical to that of IBP. The difference is in the underlying graph that it uses. IJGP typically has an accuracy parameter $i$ called i-bound, which restricts the maximum number of variables that can appear in a node (cluster). Each cluster contains a set of functions. IJGP performs message-passing on a graph called *minimal arc-labeled join-graph*.

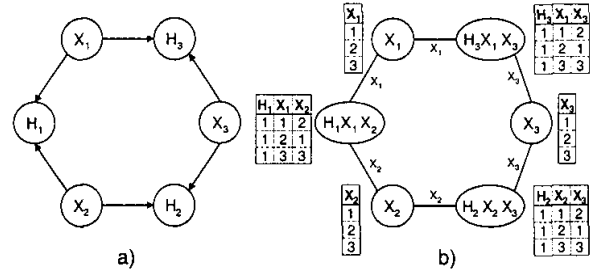

Figure 5: a) A belief network; b) An arc-minimal dual join-graph;

It is easy to define a corresponding DR-AC algorithm that operates on a similar minimal arc-label join-graph. Initially, each cluster of DR-AC can contain a number of relations, which are just the flat correspondents of the CPTs in the clusters of IJGP. The identical mechanics of the message passing ensure that all the previous results for IBP can be extended to IJGP.

### 4.2 THE INFERENCE POWER OF IBP

We will next show that the inference power of IBP is sometimes very limited and other times strong, exactly wherever arc-consistency is weak or strong.

#### 4.2.1 Cases of weak inference power

**Example 4.4** *Consider a belief network over 6 variables $X_1, X_2, X_3, H_1, H_2, H_3$ where the domain of the X variables is $\{1, 2, 3\}$ and the domain of the H variables is $\{0, 1\}$ (see Figure 5a). There are three CPTs over the scopes: $\{H_1, X_1, X_2\}$, $\{H_2, X_2, X_3\}$, and $\{H_3, X_1, X_3\}$. The values of the CPTs for every triplet of variables $\{H_k, X_i, X_j\}$ are:*

$$P(h_k = 1|x_i, x_j) = \begin{cases} 1, & if\ (3 \neq x_i \neq x_j \neq 3); \\ 1, & if\ (x_i = x_j = 3); \\ 0, & otherwise; \end{cases}$$
$$P(h_k = 0|x_i, x_j) = 1 - P(h_k = 1|x_i, x_j).$$

*Consider the evidence set $e = \{H_1 = H_2 = H_3 = 1\}$. One can see that this Bayesian network expresses the probability distribution that is concentrated in a single tuple:*

$$P(x_1, x_2, x_3|e) = \begin{cases} 1, & if\ x_1 = x_2 = x_3 = 3; \\ 0, & otherwise. \end{cases}$$

*In other words, any tuple containing an assignment of "1" or "2" for any X variable has a zero probability. The flat constraint network of the above belief network is defined over the scopes $S_1 = \{H_1, X_1, X_2\}$, $S_2 =$*

$\{H_2, X_2, X_3\}$, $S_3 = \{H_3, X_1, X_3\}$. *The constraints are defined by:* $R_{H_k, X_i, X_j} = \{(1,1,2), (1,2,1), (1,3,3), (0,1,1), (0,1,3), (0,2,2), (0,2,3), (0,3,1), (0,3,2)\}$. *Also, the prior probabilities for $X_i$'s become unary constraints equal to the full domain $\{1,2,3\}$ (assuming the priors are non-zero). An arc-minimal dual join-graph which is identical to the constraint network is given in Figure 5b.*

*In the flat constraint network, the constraints in each node are restricted after assigning the evidence values (see Figure 5b). In this case, DR-AC sends as messages the full domains of the variables and therefore no tuple is removed from any constraint. Since IBP infers the same zeros as arc-consistency, IBP will also not infer any zeros for any family or any single variable. However, since the true probability of most tuples is zero we can conclude that the inference power of IBP on this example is weak or non-existent.*

The weakness of arc-consistency as demonstrated in this example is not surprising. Arc-consistency is known to be a weak algorithm in general. It implies the same weakness for belief propagation and demonstrates that IBP is very far from completeness, at least as long as zero tuples are concerned.

The above example was constructed by taking a specific constraint network with known properties and expressing it as a belief network using a known transformation. We associate each constraint $R_S$ with a bi-valued new hidden variable $X_h$, direct arcs from the constraint variables to this new hidden variable $X_h$, and create the CPT such that:

$P(x_h = 1|x_{pa_h}) = 1$, $iff$ $x_{pa_h} \in R_S$.

while zero otherwise [Pearl1988]. The generated belief network conditioned on all the $X_h$ variables being assigned "1" expresses the same set of solutions as the constraint network.

### 4.2.2 Cases of strong inference power

The relationship between IBP and arc-consistency ensures that IBP is zero-complete whenever arc-consistency is. In general, if for a flat constraint network of a Bayesian network $\mathcal{B}$, arc-consistency removes all the inconsistent domain values (it creates minimal domains), then IBP will also discover all the true zeros of $\mathcal{B}$. We next consider several classes of constraints that are known to be tractable.

**Acyclic belief networks.** When the belief network is acyclic, namely when it has a dual join-graph that is a tree, the flat network is an acyclic constraint network that can be shown to be solvable by distributed relational arc-consistency [Dechter1992]. Note that acyclic Bayesian networks is a strict superset of polytrees. The solution requires only one iteration (two passes) of IBP. Therefore:

**Proposition 4** *IBP is complete for acyclic networks, when applied to the tree dual join-graph (and therefore it is also zero-complete).*

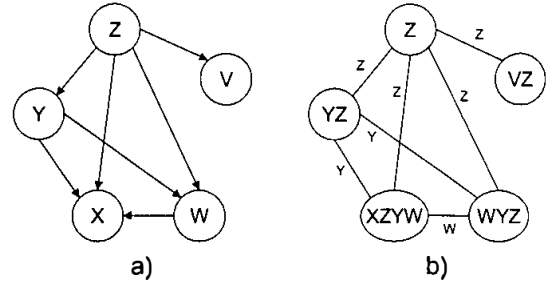

Figure 6: a) A belief network that corresponds to a Max-closed relation; b) An arc-minimal dual join-graph;

**Example 4.5** *We refer back to the example of Figure 3. The network is acyclic because there is a dual join-graph that is a tree, given in Figure 3c, and IBP will be zero-complete on it. Moreover, IBP is known to be complete in this case.*

**Belief networks with no evidence.** Another interesting case is when the belief network has no evidence. In this case, the flat network always corresponds to the *causal constraint network* defined in [Dechter and Pearl1991]. The inconsistent tuples or domain values are already explicitly described in each relation, and new zeros do not exist. Indeed, it is easy to see (either directly or through the flat network) that:

**Proposition 5** *IBP is zero-complete for any Bayesian network with no evidence.*

In fact, it can be shown [Bidyuk and Dechter2001] that IBP is also complete for non-zero posterior beliefs of many variables when there is no evidence.

**Max-closed constraints**. Consider next the class of Max-closed relations defined as follows. Given a domain $D$ that is linearly ordered let *Max* be a binary operator that returns the largest element among 2. The operator can be applied to 2 tuples by taking the pair-wise operation [Jeavons and Cooper1996].

DEFINITION **4.2 (Max-closed relations)** *A relation is Max-closed if whenever $t_1, t_2 \in R$ so is $Max(t_1, t_2)$. A constraint network is Max-closed if all its constraints are Max-closed.*

It turns out that if a constraint network is Max-closed, it can be solved by distributed arc-consistency. Namely, if no domain becomes empty by the arc-consistency algorithm, the network is consistent. While arc-consistency is not guaranteed to generate minimal domains, thus removing all inconsistent values, it can generate a solution by selecting the maximal value from the domain of each variable. Accordingly, while IBP will not necessarily discover all the zeros, all the largest non-zero values in the domains of each variable are true non-zeros.

Therefore, for a belief network whose flat network is Max-closed IBP is likely to be powerful for generating zero tuples.

**Example 4.6** *Consider the following belief network: There are 5 variables* $\{V, W, X, Y, Z\}$ *over domains* $\{1, 2, 3, 4, 5\}$. *and the following CPTs:*

$$P(x|z, y, w) \neq 0, \quad iff \ 3x + y + z \geq 5w + 1$$
$$P(w|y, z) \neq 0, \quad iff \ wz \geq 2y$$
$$P(y|z) \neq 0, \quad iff \ y \geq z + 2$$
$$P(v|z) \neq 0, \quad iff \ 3v \leq z + 1$$
$$P(Z = i) = 1/4, \quad i \in \{1, 2, 3, 4\}$$

*All the other probabilities are zero. Also, the domain of W does not include 3 and the domain z does not include 5. The problem's acyclic graph is given in Figure 6a. It is easy to see that the flat network is the set of constraints over the above specified domains:* $w \neq 3$, $z \neq 5$, $3v \leq z + 1$, $y \geq z + 2$, $3x + y + z \geq 5w + 1$, $wz \geq 2y$. *An arc-minimal dual join-graph with singleton labels is given in Figure 6b. It has 5 nodes, one for each family in the Bayesian network. If we apply distributed relational consistency we will get that the domains are:* $D_V = \{1\}$, $D_W = \{4\}$, $D_X = \{3, 4, 5\}$, $D_Y = \{4, 5\}$ *and* $D_Z = \{2, 3\}$. *Since all the constraints are Max-closed and since there is no empty domain the problem has a solution given by the maximal values in each domain:* $V = 1$, $W = 4$, $X = 5$, $Y = 5$, $Z = 3$. *The domains are not minimal however: there is no solution having* $X = 3$ *or* $X = 4$.

*Based on the correspondence with arc-consistency, we know that applying IBP to the dual join-graph will indeed infer all the zero domains except those of X, which validates that IBP is quite powerful for this example.*

An interesting case for propositional variables is the class of Horn clauses. A Horn clause can be shown to be Min-closed (by simply checking its models). If we have an acyclic graph, and we associate every family with a Horn clause expressed as a CPT in the obvious way, then applying Belief propagation on a dual join-graph can be shown to be nothing but the application of unit propagation until there is no change. It is well known that unit propagation decides the consistency of a set of Horn clauses (even if they are cyclic). However, unit propagation will not necessarily generate the minimal domains, and thus not infer all the zeros, but it is likely to behave well.

**Implicational constraints.** Finally, a class that is known to be solvable by path-consistency is implicational constraints, defined as follows:

DEFINITION 4.3 *A binary network is implicational, iff for every binary relation every value of one variable is consistent either with only one or with all the values of the other variable [Kirousis1993]. A Bayesian network is implicational if its flat constraint networks is.*

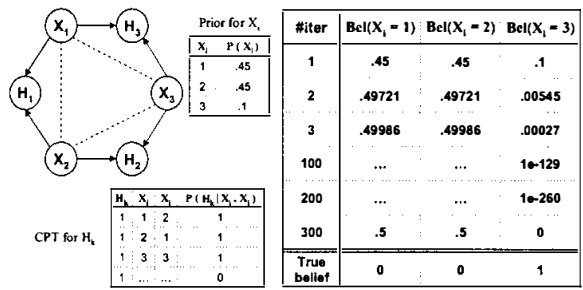

Figure 7: Example of a finite precision problem;

Clearly, a binary function is an implicational constraint. Since IBP is equivalent to arc-consistency only, we cannot conclude that IBP is zero-complete for implicational constraints. This raises the question of what corresponds to path-consistency in belief networks, a question which we do not attempt to answer at this point.

### 4.3 A FINITE PRECISION PROBLEM

Algorithms should always be implemented with care on finite precision machines. We mention here a case where IBP's messages converge in the limit (i.e. in an infinite number of iterations), but they do not stabilize in any finite number of iterations. Consider again the example in Figure 5 with the priors on $X_i$'s given in Figure 7. If all nodes $H_k$ are set to value 1, the belief for any of the $X_i$ variables as a function of iteration is given in the table in Figure 7. After about 300 iterations, the finite precision of our computer is not able to represent the value for $Bel(X_i = 3)$, and this appears to be zero, yielding the final updated belief $(.5, .5, 0)$, when in fact the true updated belief should be $(0, 0, 1)$. This does not contradict our theory, because mathematically, $Bel(X_i = 3)$ never becomes a true zero, and IBP never reaches a quiescent state.

## 5 EMPIRICAL EVALUATION

We tested the performance of IBP and IJGP both on cases of strong and weak inference power. In particular, we looked at networks where probabilities are extreme and checked if the properties of IBP with respect to zeros also extend to $\epsilon$ small beliefs.

### 5.1 ACCURACY OF IBP ACROSS BELIEF DISTRIBUTION

We investigated empirically the accuracy of IBP's prediction across the range of belief values from 0 to 1. Theoretically, zero values inferred by IBP are proved correct, and we hypothesize that this property extends to $\epsilon$ small beliefs. That is, if the flat network is easy for arc-consistency and IBP infers a posterior belief close to zero, then it is likely to be correct.

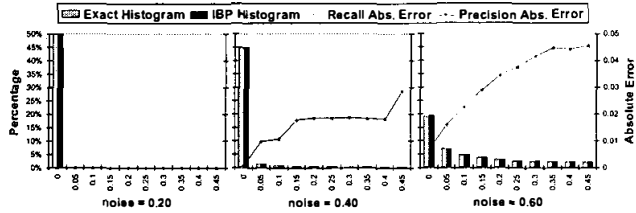

Figure 8: Coding, N=200, 1000 instances, w*=15;

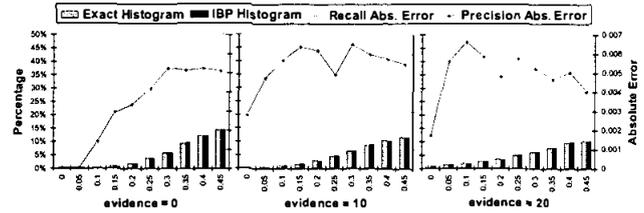

Figure 10: Random, N=80, 100 instances, w*=15;

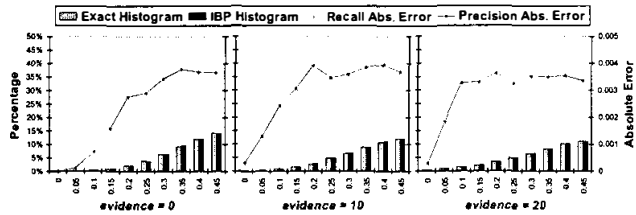

Figure 9: 10x10 grids, 100 instances, w*=15;

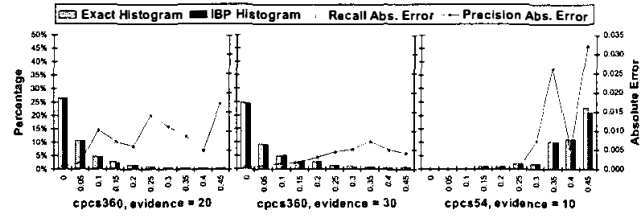

Figure 11: CPCS54, 100 instances, w*=15; CPCS360, 5 instances, w*=20;

To capture the accuracy of IBP we computed its absolute error per intervals of $[0, 1]$. Using names inspired by the well known measures in information retrieval, we use *Recall Absolute Error* and *Precision Absolute Error*. Recall is the absolute error averaged over all the exact posterior beliefs that fall into the interval. For *Precision*, the average is taken over all the approximate posterior belief values computed by IBP that fall into the interval. Our experiments show that the two measures are strongly correlated. We also show the histograms of distribution of belief for each interval, for the exact and for IBP, which are also strongly correlated. The results are given in Figures 8-11. The left Y axis corresponds to the histograms (the bars), the right Y axis corresponds to the absolute error (the lines). All problems have binary variables, so the graphs are symmetric about 0.5 and we only show the interval $[0, 0.5]$. The number of variables, number of iterations and induced width w* are reported for each graph.

**Coding networks** are the notorious case where IBP has impressive performance. The problems are from the class of linear block codes, with 50 nodes per layer and 3 parent nodes. Figure 8 shows the results for three different values of channel noise: 0.2, 0.4 and 0.6. For noise 0.2, all the beliefs computed by IBP are extreme. The Recall and Precision are very small, of the order of $10^{-11}$. So, in this case, all the beliefs are very small ($\epsilon$ small) and IBP is able to infer them correctly, resulting in almost perfect accuracy (IBP is indeed perfect in this case for the bit error rate). When the noise is increased, the Recall and Precision tend to get closer to a bell shape, indicating higher error for values close to 0.5 and smaller error for extreme values. The histograms also show that less belief values are extreme as the noise is increased, so all these factors account for an overall decrease in accuracy as the channel noise increases.

**Grid networks** results are given in Figure 9. Contrary to the case of coding networks, the histograms show higher concentration around 0.5. The absolute error peaks closer to 0 and maintains a plateau, as evidence is increased, indicating less accuracy for IBP.

**Random networks** results are given in Figure 10. The histograms are similar to those of the grids, but the absolute error has a tendency to decrease towards 0.5 as evidence increases. This may be due to the fact that the total number of nodes is smaller (80) than for grids (100), and the evidence can in many cases make the problem easier for IBP by breaking many of the loops (in the case of grids evidence has less impact in breaking the loops).

**CPCS networks** are belief networks for medicine, derived from the Computer based Patient Case Simulation system. We tested on two networks, with 54 and 360 variables. The histograms show opposing trends in the distribution of beliefs. Although irregular, the absolute error tends to increase towards 0.5 for cpcs54. For cpcs360 it is smaller around 0 and 0.5.

We note that for all these types of networks, IBP has very small absolute error for values close to zero, so it is able to infer them correctly.

### 5.2 GRAPH-COLORING TYPE PROBLEMS

We also tested the behavior of IBP and IJGP on a special class of problems which were designed to be hard for belief propagation algorithms in general, based on the fact that arc-consistency is poor on the flat network.

We consider a graph coloring problem which is a generalization of example 4.4, with $N = 20$ $X$ nodes, rather than 3, and a variable number of $H$ nodes defining the density of the constraint graph. $X$ variables are 3-valued root nodes, $H$ variables are bi-valued and each has two parents which are $X$ variables, with the CPTs defined like in example 4.4. Each $H$ CPT actually models a binary constraint between two $X$ nodes. All $H$ nodes are assigned value 1. The flat network of this kind of problems has only one solution, where every $X$ has value 3. In our experiments we also added noise to the $H$ CPTs, making probabilities $\epsilon$ and $1 - \epsilon$ rather than 0 and 1.

Table 1: Graph coloring type problems: 20 root variables

| | $\epsilon$ | H=40, w*=5 | H=60, w*=7 | H=80, w*=9 |
|---|---|---|---|---|
| | | Absolute error | | |
| IBP | 0.0 | 0.4373 | 0.4501 | 0.4115 |
| | 0.1 | 0.3683 | 0.4497 | 0.3869 |
| | 0.2 | 0.2288 | 0.4258 | 0.3832 |
| IJGP(2) | 0.0 | 0.1800 | 0.1800 | 0.1533 |
| | 0.1 | 0.3043 | 0.3694 | 0.3189 |
| | 0.2 | 0.1591 | 0.3407 | 0.3022 |
| IJGP(4) | 0.0 | 0.0000 | 0.0000 | 0.0000 |
| | 0.1 | 0.1211 | 0.0266 | 0.0133 |
| | 0.2 | 0.0528 | 0.1370 | 0.0916 |
| IJGP(6) | 0.0 | 0.0000 | 0.0000 | 0.0000 |
| | 0.1 | 0.0043 | 0.0000 | 0.0132 |
| | 0.2 | 0.0123 | 0.0616 | 0.0256 |

The results are given in Table 1. We varied parameters along two directions. One was increasing the number of $H$ nodes, corresponding to higher densities of the constraint network (the average induced width $w*$ is reported for each column). The other was increasing the noise parameter $\epsilon$. We averaged over 50 instances for each combination of these parameters. In each instance, the priors for nodes $X$ were random uniform, and the parents for each node $H$ were chosen randomly. We report the absolute error, averaged over all values, all variables and all instances. We should note that these are fairly small size networks (w*=5-9), yet they prove to be very hard for IBP and IJGP, because the flat network is hard for arc-consistency. It is interesting to note that even when $\epsilon$ is not extreme anymore (0.2) the performance is still poor, because the structure of the network is hard for arc-consistency. IJGP with higher i-bounds is good for $\epsilon = 0$ because it is able to infer some zeros in the bigger clusters, and these propagate in the network and in turn infer more zeros.

## 6 CONCLUSIONS

The paper investigates the behavior of belief propagation algorithms by making analogies to well known and understood algorithms from constraint networks. By a simple transformation, called flattening of the Bayesian network, IBP (as well as any generalized belief propagation algorithm) can be shown to work similar to distributed relational arc-consistency relative to zero tuples generation. In particular we show that IBP's inference of zero beliefs converges and is sound.

While the theoretical results presented here are straightforward, they help identify new classes of problems that are easy or hard for IBP. Based on empirical work, we observe that good performance of IBP and many small beliefs indicate that the flat network is likely to be easy for arc-consistency. On the other hand, when we generated hard networks for arc-consistency, IBP was very poor in spite of the presence of many zero beliefs. We believe that the success of IBP for coding networks can be explained by the presence of many small beliefs on one hand, and by an easy-for-arc-consistency flat network on the other.

### Acknowledgments

This work was supported in part by the NSF grant IIS-0086529 and MURI ONR award N00014-00-1-0617.